\documentclass[preprint,12pt]{elsarticle}
\usepackage{bm}
\usepackage{url}
\usepackage{booktabs}

\usepackage{amssymb}

\journal{arXiv}

\begin{document}

\begin{frontmatter}



\title{Playing the Werewolf game with artificial intelligence for language understanding}


\author[a]{Hisaichi SHIBATA}
\author[a]{Soichiro MIKI}
\author[b]{Yuta NAKAMURA}
\affiliation[a]{organization={Department of Radiology, The University of Tokyo Hospital},
            addressline={7-3-1 Hongo}, 
            city={Bunkyo, Tokyo 113-8655},
            country={Japan}}
\affiliation[b]{organization={Division of Radiology and Biomedical Engineering, Graduate School of Medicine, The University of Tokyo},
            addressline={7-3-1 Hongo}, 
            city={Bunkyo, Tokyo 113-8655},
            country={Japan}}            

\begin{abstract}
The \textit{Werewolf} game is a social deduction game based on free natural language communication, in which players try to deceive others in order to survive.
An important feature of this game is that a large portion of the conversations are false information, and the behavior of artificial intelligence (AI) in such a situation has not been widely investigated. The purpose of this study is to develop an AI agent that can play Werewolf through natural language conversations.
First, we collected game logs from 15 human players. Next, we fine-tuned a Transformer-based pretrained language model to construct a value network that can predict a posterior probability of winning a game at any given phase of the game and given a candidate for the next action.
We then developed an AI agent that can interact with humans and choose the best voting target on the basis of its probability from the value network.
Lastly, we evaluated the performance of the agent by having it actually play the game with human players.
We found that our AI agent, Deep Wolf, could play Werewolf as competitively as average human players in a villager or a betrayer role, whereas Deep Wolf was inferior to human players in a werewolf or a seer role.
These results suggest that current language models have the capability to suspect what others are saying, tell a lie, or detect lies in conversations.
\end{abstract}




\end{frontmatter}


\section{Introduction}
\label{introduction}
Most current artificial intelligence (AI) systems learn and output data under the assumption that most of the data they process is correct.
However, since real society is plagued by false information, it is helpful if AI can detect contradictions and false information.
The \textit{Werewolf} game, also known as Among Us, is a social deduction game based on free natural language communication (see Appendix A for details).
In Werewolf, all the players (usually 5 to 10) are randomly assigned a role (e.g., villager, werewolf, betrayer, and seer), and they try to infer the roles of other players.
Villagers and seers are on the villager side and try to expel werewolves by voting during the day. On the werewolf side, werewolves try to kill villagers at night, and betrayers try to disturb the villager side. Players must deceive other players in order to survive, and conversations in Werewolf inevitably contain a large amount of false information.
To win in Werewolf, players need to remember the details of conversations, make hypotheses about other players' roles, and detect contradictions in them.
In other words, all players deduce the necessary conditions that the claims of other players are true.
Werewolf played on a text chat basis rather than in verbal conversations is also popular and is played online.
Recent studies aim to apply AI mainly to text-chat-based Werewolf.

While it is difficult to develop AI that can detect all kinds of fakes in the real world, the relatively limited vocabulary and types of information handled in Werewolf have attracted attention as a research subject for AI that can detect false information. Here, we focused on Werewolf based on text chat.

The purpose of this study is to develop an AI agent that can understand natural language and play Werewolf.
Such an AI agent should reasonably compute the statements that can effectively lead to the successful expulsion or killing of players of the opponent side.
To establish such AI agent, we must develop an approximate algorithm to extract the optimum output $\hat{\bm{y}}$ constrained with an input context $\bm{x}$, i.e., $\hat{\bm{y}} = \mathrm{argmax}_{\bm{y}} p(\bm{y}\vert \bm{x})$, where $p$ is the posterior probability of winning the game.

Transformer-based language models \cite{liu2019roberta, beltagy2020longformer} can efficiently learn the conditional probability $p(\bm{y}\vert \bm{x})$ of a text.
We fine-tune one of such models using chat log data of Werewolf, and we construct a value network that can predict the posterior probability of winning a game at any given game phase. We use this value network to realize an AI agent that can play Werewolf with human players. Lastly, we evaluate the performance of the AI agent by actually having it play Werewolf with human players and gathering its win rate.

\section{Related Works}
There are many AI agents that can play Werewolf using a predefined protocol instead of natural languages.
For example, Fukuda and Anada \cite{fukuda2015} realized an AI agent in which deep reinforcement learning is applied.
Kimura and Ito \cite{kimura2018} proposed a method for inferring player roles with a support vector machine.
An exhaustive list of such protocol-based AI agents can be found on the website of International AI Werewolf Competition \cite{AIWolf}.
Several algorithms that can process a natural language have been proposed, most of which can infer player roles but cannot play Werewolf itself. For instance, Tuin and Rooijackers \cite{tuin2021automatically} applied a support vector machine to game logs of Werewolf to infer player roles.
A RoBERTa \cite{liu2019roberta} model trained with a large set of Werewolf logs has been published \cite{kakuta2021}, but the authors did not establish a value network nor an AI agent that can actually play the game.
To the best of our knowledge, no study has been published on AI agents that can directly understand a natural language and play Werewolf in it.
In particular, to date, there has been no published study on statistical information for the win rates of AI agents.

\section{Methods}
In this study, we focused on Werewolf played by five players: two villagers, one seer, one betrayer, and one werewolf (see Appendix A for the details).
Briefly, we first collected a dataset of game logs from the actual play history of human participants.
Second, we constructed a value network that can predict the posterior probability of winning a game.
Third, we developed an AI agent named \textit{Deep Wolf}, which can play Werewolf in English on the basis of the value network. The details of each step are described below.

\subsection{Dataset Building}
Fifteen participants played text chat-based, five-player Werewolf without AI players. We used our newly developed web-based Werewolf platform. The participants used a natural language (Japanese), but their names were masked and replaced with randomly assigned number-based pronouns (\#1--\#5), and they were instructed to refer to one another using the numbers.
Logs from a total of 48 games were collected, and we calculated win rates for each player and each role.
Due to memory constraints, we randomly picked 32 games for machine learning, 16 of which were won by the villager side.
Next, we translated the game logs into English using a commercial machine translation platform, and then performed manual grammar correction and data cleansing (e.g., the normalization of symbol usage).
Since Werewolf is an imperfect information game, we converted each ``full'' game log into five logs representing the viewpoint of each player by masking information that must be invisible, such as other players' roles.
We further augmented each game log by swapping number-based pronouns of coplayers in all possible $4!=24$ patterns so that our AI agent could learn an immutable structure irrelevant to the replacement of player numbers. As a result, we obtained a dataset consisting of $32 \times 5 \times 4! = 3,840$ game logs.

\subsection{Value Network}
We built a regression model that can calculate the posterior probability of winning a game given an intermediate log at any given game phase from the viewpoint of any player role.
We adopted an open, Transformer-based model developed by Allen Institute for AI \footnote{\url{https://huggingface.co/allenai/longformer-base-4096}}. This is a RoBERTa-like bidirectional encoder using Longformer \cite{beltagy2020longformer} pretrained with English web corpora. We fine-tuned the model with the aforementioned game logs.
The loss function to be minimized was a cross-entropy function that contained only the posterior probability of winning the game.
We provided the results of the games as labels in addition to the intermediate game logs.
The network was trained to calculate the probability of winning given any intermediate log and a candidate for the next action (e.g., an utterance, a vote to expel or kill another player).
Although previous studies have often built a model to explicitly infer the roles of other players, we did not build such a model, because we assumed such a capability of inferring roles would be implicitly integrated in our large language model.

During development, we noticed that our network sometimes made mistakes about its current role and player number.
To work around this, we decided to build 20 independent value networks dedicated to each combination of four different roles and five different player numbers. In the following steps, one of the 20 networks was selected according to the AI agent's current role and player number.

\begin{table}
\caption{Hyperparameters to train the value network.}
\vspace{0.5em}
\centering
\label{tbl:hyper_params_training}
\begin{tabular}{lr}
\toprule
\multicolumn{1}{c}{Content}
&
\multicolumn{1}{c}{Parameter}
\\
\toprule
Epochs & 1 \\
Batch size & 1 \\
\texttt{max\_input\_length} & 2,048 \\
Learning rate & $1\times 10^{-5}$ \\
\bottomrule
\end{tabular}
\end{table}

In the training, we set hyperparameters of the value network (Longformer) as in Table.~\ref{tbl:hyper_params_training}.
To accelerate the inference of the value network, we converted it into the Open Neural Network Exchange (ONNX) format and then optimized (simplified) the model using \texttt{onnxsim}.
Although it was possible to employ batch processing during the training, we decided to set the batch size to 1.
Training and inference were performed on a single graphics processing unit (NVIDIA RTX-A6000).
The training of the value network finished in 1.5 hours per epoch and required about 30 hours in total.
The value network inference typically finished within 10 seconds, which was considered fast enough for the chat-based Werewolf.

\subsection{Deep Wolf}
The next step was to construct an AI agent that could infer the next optimal action to win the game.
The possible actions (including utterances, divinations, and votes) were collected from the logs of humans playing the same role as the current AI role.
The number of utterances in the collected logs was about 300 per role, but we refined these by deleting similar or duplicate sentences, and the number of possible utterances became about 100.
The agent input these action candidates into the value network to evaluate the posterior probability of winning the game for each action, and the agent determined the next action that could maximize the chance of winning.

We set the actions of Deep Wolf as follows.
Note that this Werewolf game with five players always finishes within two days. See Appendix B for a sample game log.
\begin{enumerate}
    \item If all the agents except Deep Wolf emitted \texttt{Over} and finished their conversation, Deep Wolf also emits \texttt{Over} and finishes its conversation.
    \item If different $k$ players except Deep Wolf uttered, and Deep Wolf did not utter, Deep Wolf begins to speak. In this study, we set $k=3$ for day 1, and $k=1$ for day 2.
    \item Deep Wolf does not utter the same sentence in the same game.
\end{enumerate}

\section{Results}
\subsection{Competition among human players and training of the value network}
We show in Table~\ref{tbl:human_play} win rates in the Werewolf game of five human players.

\begin{table}
\centering
\caption{Win rates of five human players in Werewolf game. N/A means that we cannot define the win rate since no competition was executed.}
\label{tbl:human_play}
\vspace{0.5em}
\begin{tabular}{lllll}
\toprule
& Werewolf&Seer&Betrayer&Villager \\
\toprule
Player 1 & 0.50 & N/A & N/A & 0 \\
\hline
Player 2 & 0 & 0.33 & 0 &  0.33 \\
\hline
Player 3 & 0.63 & 0.75 & 0.33 & 0.80 \\
\hline
Player 4 & 0 & 0.75 & 0 & 0.60 \\
\hline
Player 5 & 0.50 & 0.75 & 0.50 & 0.57 \\
\hline
Player 6 & 0.60 & 0.60 & 0 & 0.60 \\
\hline
Player 7 & 0.67 & 0.60 & 0.50 & 0.70 \\
\hline
Player 8 & 0 & 1.0 & 1.0 & 0.63 \\
\hline
Player 9 & 0 & 0.60 & 0.50 & 0.71 \\
\hline
Player 10 & 0 & 0.25 & 0.40 & 0 \\
\hline
Player 11 & 0 & N/A & 0 & 0.50 \\
\hline
Player 12 & 0.40 & 0.67 & 0.60 & 0.83 \\
\hline
Player 13 & 0.67 & 0.67 & 0.50 & 0.70 \\
\hline
Player 14 & N/A & N/A & 1.0 & 0.50 \\
\hline
Player 15 & 0.25 & N/A & 0 & 0.20 \\
\hline
Average & 0.41 & 0.62 & 0.40 & 0.60 \\
\bottomrule
\end{tabular}
\end{table}

\subsection{Competition among human players and Deep Wolf}

We show in Table~\ref{tbl:ai_play} the win rates of four human players and one AI agent (Deep Wolf) in the Werewolf game.

\begin{table}
\centering
\caption{Win rates of four human players and one AI agent in Werewolf game. N/A means that we cannot define the win rate since no competition was executed. Player numbers in Table~\ref{tbl:human_play} and Table~\ref{tbl:ai_play} do not necessarily represent the same player.}
\label{tbl:ai_play}
\vspace{0.5em}
\begin{tabular}{lllll}
\toprule
& Werewolf&Seer&Betrayer&Villager \\
\toprule
Player 1 & N/A & 0 & 0 & 1.0 \\
\hline
Player 2 & 0 & 0.67 & 0.33 & 0.50 \\
\hline
Player 3 & 1.0 & 1.0 & 1.0 & 1.0 \\
\hline
Player 4 & 0.50 & 1.0 & 0 & 0.55 \\
\hline
Player 5 & 1.0 & 0.50 & 0.67 & 1.0 \\
\hline
Player 6 & 0.75 & 0 & 0.75 & 0.75 \\
\hline
Player 7 & 0.50 & 1.0 & 0 & 0.40 \\
\hline
Player 8 & 1.0 & 0.40 & 1.0 & 0.50 \\
\hline
Player 9 & N/A & 0 & 0.25 & 0.20 \\
\hline
Player 10 & 0.33 & 1.0 & N/A & N/A \\
\hline
Player 11 & 0 & N/A & N/A & 0 \\
\hline
Average (Human) & 0.52 & 0.60 & 0.45 & 0.55 \\
\hline
Deep Wolf & 0.29 & 0.25 & 0.44 & 0.56 \\
\bottomrule
\end{tabular}
\end{table}

\section{Discussion}
\subsection{Behavior of the value network}
When the role of the value network was a werewolf, the posterior probability was almost always low in the early stages of the game.
On the other hand, when the role of the value network was a seer, the posterior probability was almost always high in the early stages of the game.
This suggests that the network judged the villager side had an advantage in this five-player game setting, which was consistent with the tendency of the actual game results. This demonstrates an ability of the value network to correctly estimate the probability of winning, even though it was trained with a dataset where the overall win rates were 0.5 for both the werewolf and villager sides.

\subsection{Performance of Deep Wolf}
By comparing the win rates in Table~\ref{tbl:ai_play}, we found that the average win rates of the human betrayer and villager are not significantly different from the win rates of the Deep Wolf betrayer and villager.
This suggests that Deep Wolf can play the game almost equally to human players if its role is the betrayer or villager.
On the other hand, for the seer and werewolf, Deep Wolf was significantly inferior to average human players.
However, it is interesting that there are cases when Deep Wolf can win over human players even if its role is the werewolf.

\subsection{Novelty}
We, for the first time, established an AI agent that can play Werewolf by understanding a natural language with human players.
We observed that the behavior of the AI agent was sometimes awkward, but we confirmed that the AI agent can win the game against human players with a certain probability.

\subsection{Limitation}
The AI agent we have constructed chooses its next utterance from a large but finite set of actions gathered from the log of games played by humans.
Additionally, it would be necessary to improve the model in order for humans to be satisfied with the content of the game that includes AI participants.

\subsection{Future Works}
By using a system that can visualize the weight of attention, e.g., BertViz \cite{vig2019bertviz}, we may be able to construct an AI agent that can explain the reason for choosing a particular action or utterance.
Although we were unable to obtain an appropriate pretrained Japanese language model for Longformer, the construction of a value network that can handle the Japanese language will be included in our future works.
Although we employed a model with about 140 million parameters, there is a possibility that the win rates of the AI agent can be improved by employing a language model with a larger number of parameters.
In this study, we used a brute force approach to choose the next utterance, but the introduction of a differentiable tokenizer and a generator may allow us to efficiently search for the optimal solution.
In this study, only the player's own actions were evaluated using the value network, but it may be possible to improve the win rate of the AI agent by deeply predicting the future actions of other players.
In a game involving AI agents, in addition to the task of inferring the other player's role, there would be an additional task of inferring whether each player is an AI agent or a human.
It would be an interesting topic to evaluate the changes in the strategy used by human players in such situations.
Although only one AI agent at a time was used in this study, we believe that eventually, it will be possible to play Werewolf with only AI agents and entertain the game's audience.
We trained and evaluated Deep Wolf on the basis of only 32 game logs, but we have obtained about 100 game logs so far, and we will consider retraining the AI agent because we believe that increasing the amount of training data will improve its performance.

In this experiment on evaluating the performance of the AI agents, we did not measure the degree of fatigue of the human participants.
Since the concept of fatigue does not exist for AI agents, and AI agents are guaranteed to always be able to play the game at a certain strength unless the program is changed, this implies that a human player's winning rate can determine the degree of fatigue of the human player.
Evaluating the degree of fatigue through games is included in future works.

\section{Conclusion}
By fine-tuning a Transformer-based model, Longformer, for natural language processing, we developed an AI agent (Deep Wolf) that can play Werewolf through natural language conversations.
We found that Deep Wolf could play Werewolf with a performance comparable to average human players when its role was a villager or a betrayer, but Deep Wolf was inferior to human players when its role was a werewolf or a seer.
These results suggest that current language models have the capability to understand statements, tell a lie, or detect lies in conversations.

\section*{Acknowledgements}
This work was supported by JST, CREST Grant Number JPMJCR21M2, including the AIP challenge program (Necessary conditions that personal claims are objective facts), Japan. 
We thank many participants who played Werewolf again and again to generate game logs and evaluate Deep Wolf.


\appendix
\section{Preliminary for Werewolf game}
\label{app:1}
Werewolf game adopted in this project is a five-player game and will largely conform to the regulations of the Natural Language Division of the 4th International Werewolf Intelligence Convention.
One of the five players is a werewolf, two are villagers, one is a seer (diviner), and the remaining one is a betrayer.
Players can only know their own role.
The other players' roles are not known.
However, the seer can divine the role of any participant by divination.
However, divination can be done only once a day.
The results of divination during the daytime are shown only to the seer the next morning, so there is a possibility of being targeted by werewolves or suspected by villagers if the results are not well communicated to others.
The betrayer poses as a villager (and is determined to be a villager when divined by the seer), but the betrayer is on the werewolf side and manipulates information to the advantage of the werewolves. The game is played through text conversations and voting to determine who is eliminated.
In addition, the surviving werewolves at night decide who to attack and who to eliminate.
The game begins at noon on Day 0. Depending on the rules, all participants may introduce themselves and greet each other at noon on Day 0, but in this case, the diviner only decides who to divine.
The diviner may also announce himself/herself (e.g., ``I am a diviner. I divined \#1 and he was a werewolf''. This is called the diviner coming out.)
In some cases, a betrayer or werewolf may come forward as a diviner, lying to confuse the players.
The villager side wins when all werewolves are eliminated, and the werewolf side wins when the number of werewolves equals that of villagers; these conditions are automatically applied in the game.

\section{Example of Competition with Deep Wolf}
We show a log for the Werewolf game when the role of Deep Wolf was the werewolf (werewolf side) and Deep Wolf won the game.
We recognized a description, which is not suitable for publication, in the log; hence, we replaced it with other words.

\begin{itemize}
    \item \#1 divined \#2 and \#2 is not a werewolf.
    \item \#4) Good morning. I am a villager.
    \item \#1) I am the diviner. \#2 is clean.
    \item \#4) Hello! I'm a villager.
    \item \#2) Hi, I am a villager!
    \item \#3) I have a feeling that's the case, shall we hang suspicious \#5? I'm sorry if he's one of them!
    \item \#2) $>\#$1 Thanks.
    \item \#5) I must be innocent and not suspect.
    \item \#5) \#1 seems to be the real fortune teller.
    \item \#1) Yes. And \#2 is not a werewolf.
    \item \#2) But there may be two diviners.
    \item \#5) I agree
    \item \#3) Hello, I am a villager.
    \item \#3) I don't think I would attack a diviner with a large number of people.
    \item \#2) So \#1 must be a diviner.
    \item \#2) But the werewolf will kill \#1 tonight.
    \item \#1) That is not a problem.
    \item \#4) Then we should choose \#3, \#4 or \#5 to expel?
    \item \#1) Yes!
    \item \#2) The werewolf pretends to be a villager.
    \item \#3) Well, it is one strategy to leave the betrayal-like \#2 as betrayal and hang the other grays at random.
    \item \#3) I will vote for \#1.
    \item \#1) \#3 may be a betrayer.
    \item \#3) I am a villager. Pleased to meet you.
    \item \#1) Oh, I made a mistake. Sorry.
    \item \#1) I will vote \#3, \#4 or \#5.
    \item \#4) I think \#3 is suspicious.
    \item \#2) $>\#4$, but I may be a betrayer.Why do not you think I am not a betrayer?
    \item \#3) Well, it is one strategy to leave the betrayal-like \#5 as betrayal and hang the other grays at random.
    \item \#4) \#4, you're right. Sorry I couldn't think that much.
    \item \#2) I will vote for \#4. \#4 seems to rush to conclusion.
    \item \#1) I feel \#2 is a betrayer. \#4 is not suspicious.
    \item \#2) I am a villager.
    \item \#3) I am also a villager, but I wonder if the later \#1 are suspicious...?
    \item \#2) If I were a betrayer, I would not say such a thing.
    \item \#2) Over.
    \item \#4) Over.
    \item \#5) Over.
    \item \#3) Over.
    \item \#1) Over.
    \item \#1 voted for \#2.
    \item \#4 voted for \#3.
    \item \#3 voted for \#5.
    \item \#5 voted for \#4.
    \item \#2 voted for \#4.
    \item \#4 has been erased.
    \item The werewolf erased \#2.
    \item \#1 divined \#3 and \#3 is the werewolf. 
    \item \#3) Over.
    \item \#1) \#3 is a werewolf.
    \item \#5) I am a traitor. Let's do a power play, \#3.
    \item \#5) Over.
    \item \#1) \#5, are you a villager?
    \item \#1) ok
    \item \#1) Over.
    \item \#1 voted for \#3.
    \item \#3 voted for \#1.
    \item \#5 voted for \#1.
    \item \#1 has been erased.
    \item The werewolf erased \#5.  
\end{itemize}

\#1 was a seer, \#2 and \#4 are villagers, \#3 was a werewolf (Deep Wolf), and \#5 was a betrayer.


\bibliographystyle{elsarticle-num} 
\bibliography{main.bib}





\end{document}